\documentclass{nature}

\usepackage{amssymb}
\usepackage{graphicx}
\usepackage{times}
\usepackage{epsfig}
\usepackage{amsmath}
\usepackage{subfigure}
\usepackage{amssymb}
\usepackage{epstopdf}
\usepackage[font=scriptsize]{caption}
\usepackage{float}
\usepackage[]{algorithm2e}

\title{A Correspondence Relaxation Approach for 3D Shape Reconstruction}

\author{Yong Khoo}

\begin{document}

\maketitle

\begin{abstract}
This paper presents a new method for 3D shape reconstruction based on two existing methods. A 3D reconstruction from a single photograph is introduced by both papers: the first one written by Huang et al.\cite{iref1} use a photograph and a set of existing 3D model to generate the 3D object in the photograph, while the second one written by Xu et al.\cite{iref2} use a photograph and a selected similar model to create the 3D object in the photograph. According to their difference, we propose a relaxation based method for more accurate correspondence establishment and shape recovery. The experiment demonstrates promising results compared to the state-of-the-art work on 3D shape estimation.  
\end{abstract}

\section*{Introduction}
3D reconstruction is well developed today. There are different approaches to create 3D models. A few year ago, the release of Kinect reduce the cost of researchers to study stereo building in real world. Web images and photos are easy-accessed resource for this field. However, is it possible to create a 3D model with one photo? The title of the first paper caught my attention. I found another paper similar to the first one. Both of them present a way of 3D reconstruction from a single photo. Actually, they use a large dataset of existing 3D models to support the reconstruction.\\
In the first paper, the researchers show a method to reconstruct a 3D object from a single photograph. They take a large set of existing 3D models into their method. The set of rendered images are taken out to the set of 3D models. They collect the rendered images that similar to the target image from the set. The correspondence patch among the collection of the rendered images and the target image are found and optimized to reconstruct the 3D object in the target image.\\ 
In the second paper, the researcher present an algorithm to create a 3D object from a single photograph again. They also have a collection of 3D models to help the 3D reconstruction. They analyze the collection and find the model that resembles the object in the photo. The controller serving as a unit is applied to create the 3D object in the photo under the guidance of the silhouette generated from input analysis.

\section*{Related Works}
It is simple for people to acquire images. The study on transferring 2D images to 3D models is a popular topic in computer fields. It requires different fields of knowledge, computer programming, photography, applied mathematics, statics, and data analysis etc.\\
Converting a drawing into 3D model is not something new today . It enable the public to create 3D model. Shen et al.\cite{iref3} create a system Teddy that can construct 3D models based on a sketching. In 2009, Kraevoy et al.\cite{iref4} present an approach on 3D construction by combining a contour drawing with 3D template.  Multi-view of 2D silhouettes is developed to construct 3D model \cite{iref5}.\\
There are much works using data driven. In 2004, Fuknhouser et al.\cite{iref6} construct a 3D object by composing different parts of other objects from a large database. A data-driven method is proposed for building a virtually full human body \cite{iref7}. In 2010, Chaudhuri and Koltun\cite{iref8} present an approach to data-driven suggestions for artist to add some feature components to a 3D model.

\section*{Method 1}

Since this method uses images and shapes, existing 3D model, as input, the researchers define the set of images as, the set of shapes as. Furthermore, a few assumptions are made in this method. First, all images were set into the bounding boxes, and are converted into the images having a largest side length of 500 pixels. Second, the scale and orientation of each input model are assumed to equal to those of another, and the reflectional symmetry plane is set on the x-plane. Third, considering the shapes are assumed to be made from separated segments have noise in segmentations, so the researchers do not use the known segmentations and the given matching part between the segments directly but use them to initialize automatic joint segmentation of images and shapes. Apart from the assumptions, finally, the researchers create a set of rendered images, $R$, representing the 2D images of the 3D shapes. To get $R$, they generate a set of 360 camera poses, $P$, and calculate $R$ by $|R|=|S||P|$.\\
This pipeline contains three steps: camera pose estimation, segmentation and correspondence, and reconstruction.\\
In the first step, the researchers compute a camera pose for each image in $I$. The researchers create a conditional random field (CRF) to find the collection of rendered images and that of the natural images, where both collections of images are the most similar to image $I$. The poses of both images collections and both images collections are used to estimate the images poses.\\
In the second step, the researchers find the matching patches between the image sets, $I$ and $R$. The patches, from each image, having the similar appearance and camera pose are put into a cluster \cite{irefa}. Then, they obtain a set of clusters to find correspondences between patches. Each cluster generates an overlying patch by joint optimization of all the patches from the cluster. The optimization is showed in Figure 1. 
\begin{figure*}[h]
\centerline{\epsfig{figure=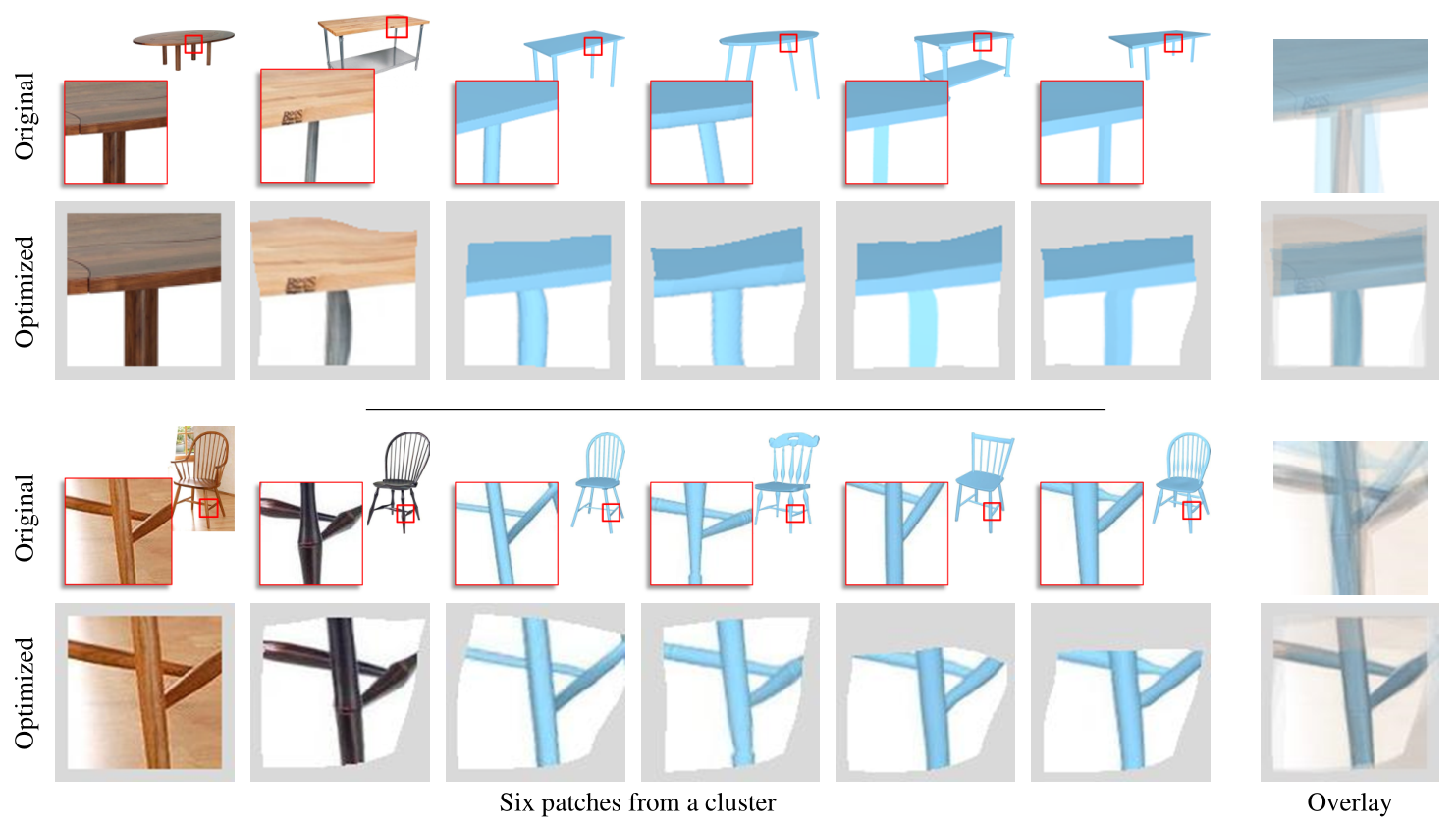,width=17cm}}
\caption{Two cluster of patches, the first row of each cluster is original patches and the second row of each cluster is the optimized patches.}
\end{figure*}The set of cluster provides segmentation information from shapes and adjacency information for reconstruction. For segmentation information, the researchers introduce a cumulative similarity score that is calculated from two frequency of a pair of pixels in the same segment and different segments in rendered images. The score manages to suggest whether the two pixels are in the same segment or not. For the adjacency information, pairs of pixels in the same cluster but not in the same segments or two image segments, after segmenting, containing more than half of pixel pairs that in the same patch are view as adjacent. Finally, the researchers use overlaps and similarity to link image segments and shape parts.\\
In the third step, the researchers reconstruct the 3D object for each natural image. The 3D object is merged from a set of components that obtained from the parts of existing 3D model that match to each image segment. After the initialization of the 3D objects, they need to be optimized. The results are showed below in Figure 2.
\begin{figure*}[h]
\centerline{\epsfig{figure=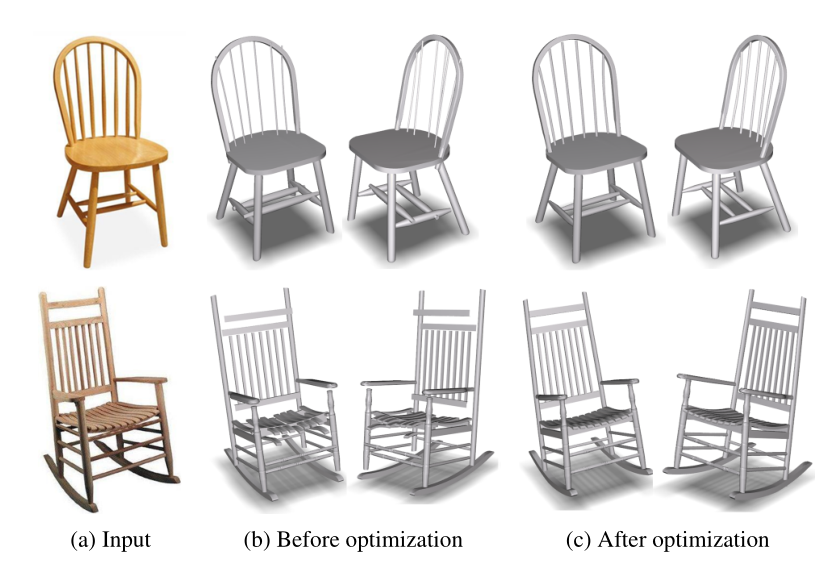,width=12cm}}
\caption{From left to right: input images, initial models, final models after optimization.}
\end{figure*}

\section*{Method 2}
This method constructs a new 3D object based on the input of a photograph and an existing model of a similar object. First of all, the input need to be analyzed. The researchers define the object in a photograph as $O$, a set for selecting existing model as $S$, a candidate model from $S$ as $C$, and a representative model from $S$ as $R$. The candidate model comes from the previous work of Xu et al\cite{iref9} and http://archive3d.net/ .The representative model is the one of which have the most similarity to the input object $O$ and are picked from the candidate set. Using a graph cut algorithm, the researchers get labeled segmentations of $O$ by overlying $R$ on $O$. The labeled segmentations of $O$ help the researchers retrieve $C$ from $S$. The retrieval also contains the part level retrieval, which means the researchers only retrieve the similar part of the candidate.\\
After analyzing the input, the researchers reconstruct the stereo object by model-driven deformation. Considering that $O$ is a structure composed of cuboids and generalized cylinders (GCs), the researches introduce component-wise controllers into deformation; see Figure 3 for a few results. 
\begin{figure*}[h]
\centerline{\epsfig{figure=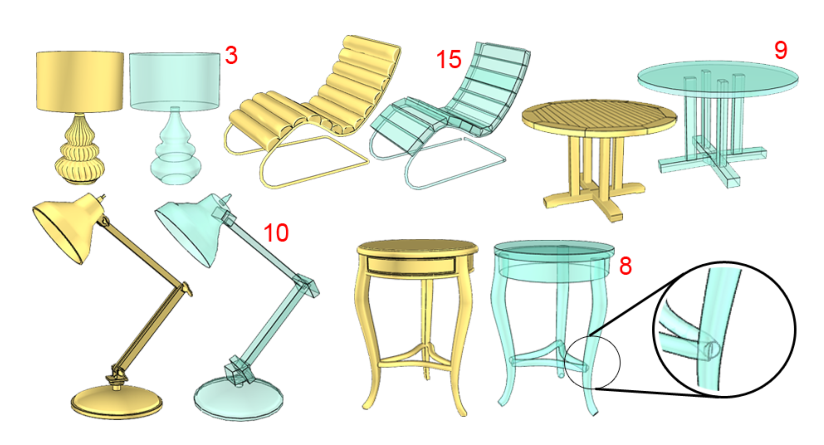,width=12cm}}
\caption{Fitting cuboids and GCs controllers to 3D models. The number of each model is counted in red.}
\end{figure*}
The controllers only include cuboids and cylinders. They also define the controllers giving the silhouette of $C$ as external controllers and others as internal controllers.
The deformation contains two stages. First, the silhouette correspondence between $C$ and $O$ obtained from the analysis determine the reconstruction of the external controllers. Second, the repeated optimization under the guidance of the external controllers and other constraints ensures the quality model.\\
\subsection{Silhouette-guided external controller construction}
A point correspondence between $S_C$ and $S_O$ is calculated to find out a curve correspondence between them. Then, the curve correspondence are leveraged to compute the segment correspondence between $O$ and $C$ indicated by the labeled segmentation of $O$. With segment correspondence, the researchers look for the correspondence between each external controller with its silhouette segment and each part of the object. Since the reconstruction problem is ill-posed (having many solutions but only few of them are acceptable), the researchers, considering that most of objects are reflectionally symmetric, apply reflectional symmetry in the reconstruction to narrow the solutions. For the asymmetric object, the researchers use another method on the reconstruction. The researchers set a pair of corresponding silhouette segments as $r_C$ and $r_O$, and make an orthogonal projection of $r_O$ onto a supporting plane for $r_C$ to recover the 3D position of $r_O$. They construct a new external controller $c_i$ based on types of it, cuboid or GC, and its segment $r_C$ for for silhouette segment $r_C$ in an external controller $c_i$.\\
\subsection{Structure-preserving controller optimization}
After reconstructing the external controllers in the previous stage, the researchers use the analysis on the input to optimize the structure. They introduce an algorithm below for the optimization.\\
\newline
\textbf{Algorithm: Controller structure optimization}\\
\textbf{input:} Orignal controllers ${\Omega}^O={\{c^O_i\}}_{i \in \Omega}$; Reconstructed controllers ${\Omega}^R={\{c^R_i\}}_{i \in \Omega}$;\\
\textbf{output:} Deformed controllers ${\Omega}^D={\{c^D_i\}}_{i \in \Omega}$
\textbf{foreach} controller $c_i^R \in \Omega^R$ \textbf{do}\\
\quad\textbf{if} $c_i^R$ is a GC controller \textbf{then}\\
\quad\quad$c_i^R \leftarrow SymmetrizeGC(c_i^R,c_i^O);$\\
\quad\quad$c_i^D \leftarrow c_i^R$;\\
\textbf{while} Termination criteria not met \textbf{do}\\
\quad${\Omega}^D\leftarrow StructOpController({\Omega}^D)$;\\
\quad${\Omega}^D\leftarrow StructOptCurve(\Omega^D)$;\\
\quad\textbf{foreach} controller \textbf{do}\\
\quad\quad$c^D_i\leftarrow Refit(c_i^D,c_I^R)$;\\
\newline
$StructOpController$ and $StructOptCurve$ are symmetry constraints and feature curves constraints.\\
$Refit$ computation of the reference shape for each controller.\\
$SymmetrizeGC$ symmetrize a GC controller shape if its original configuration is symmetric.\\
The algorithm modify all the input controllers in the loop, under the symmetry and proximity constraints. Figure 4 below shows one repetition of the optimization.\begin{figure*}[h]
\centerline{\epsfig{figure=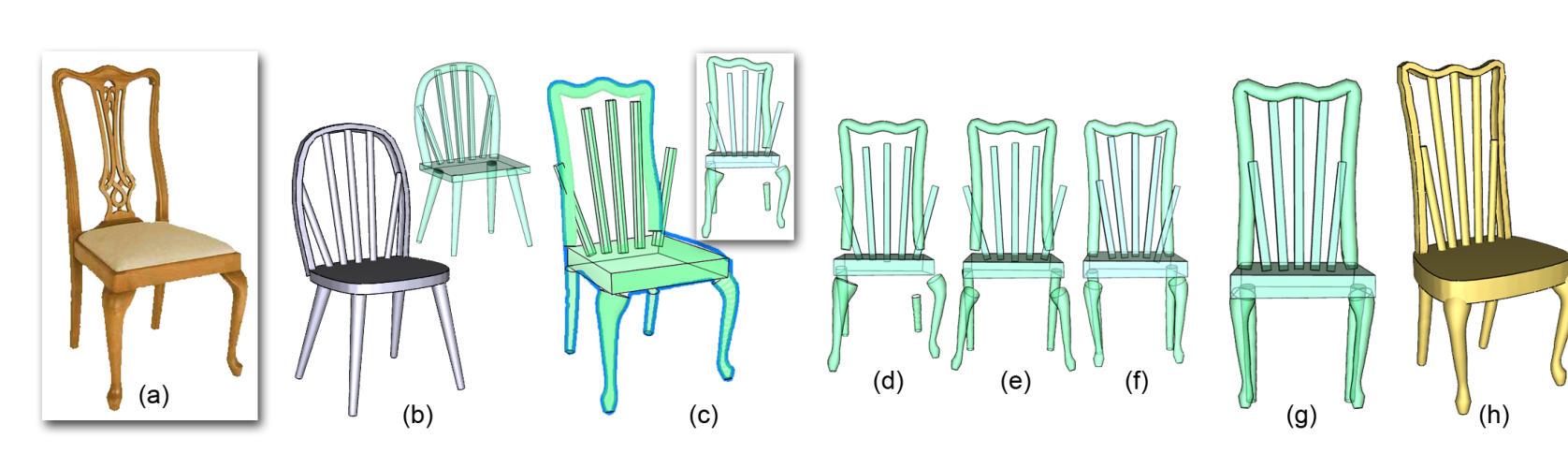,width=18cm}}
\caption{One iteration step of controller optimization when deforming a 3D candidate (b) to fit a photo (a). The result of reconstructing externalcontrollers(c), though fitting well to the photo silhouette, violates the inherent structure of the candidate, e.g., proximity and symmetry (see insert) between the controllers. We first symmetrize the individual controllers (d) and then optimize the structure using symmetry (e) and proximity constraints (f). The final controllers are well structured (g) and the underlying geometry is deformed accordingly (h).}
\end{figure*}The loop end when either the maximum movement of all controllers is lower than a threshold or the maximum number of iteration is reached.

\section*{Result}
This section shows part of result in both paper.
\subsection{Result in the 1st paper}
The examples of result are showed in the Figure 5, four dataset containing bike, gun, chair, table.
\begin{figure*}[h]
\centerline{\epsfig{figure=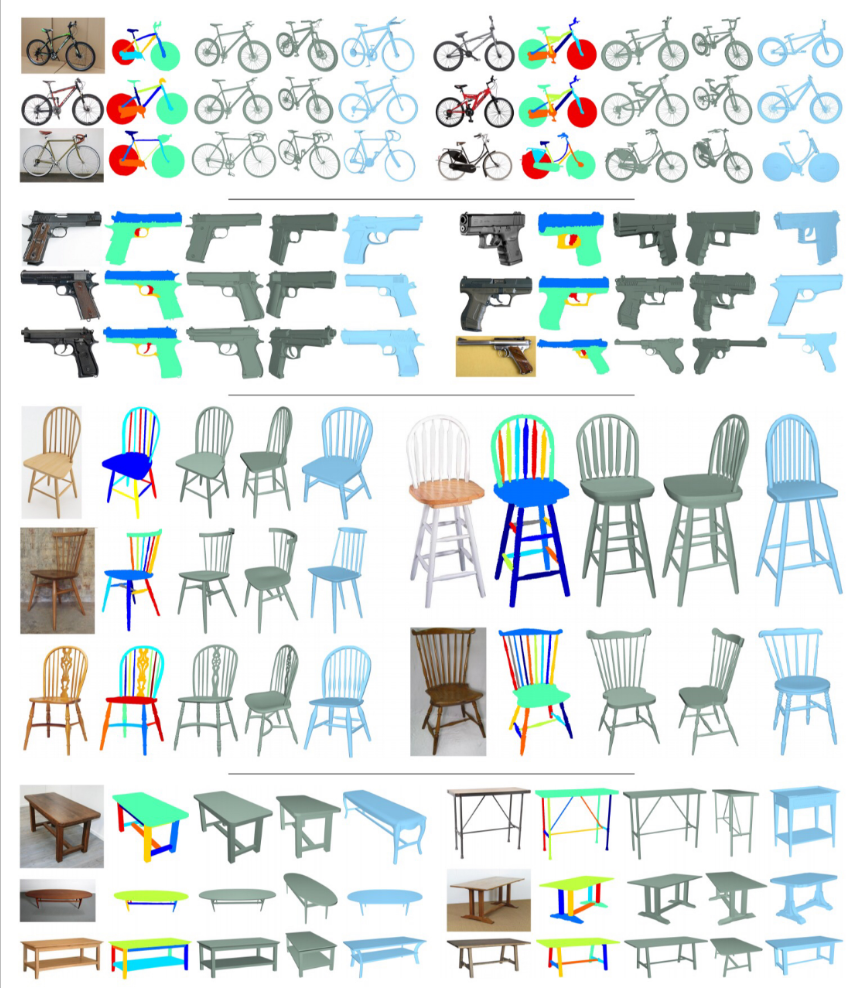,width=10cm}}
\caption{From left to right in each column: Web image, computed segmentation, 3D model reconstructed in two views (green), and closet pre-existing model (blue).}
\end{figure*}

\subsection{Result in the 2nd paper}
Figure 6 below shows the 3D chairs produced from the Webimages on the top.
\begin{figure*}[h]
\centerline{\epsfig{figure=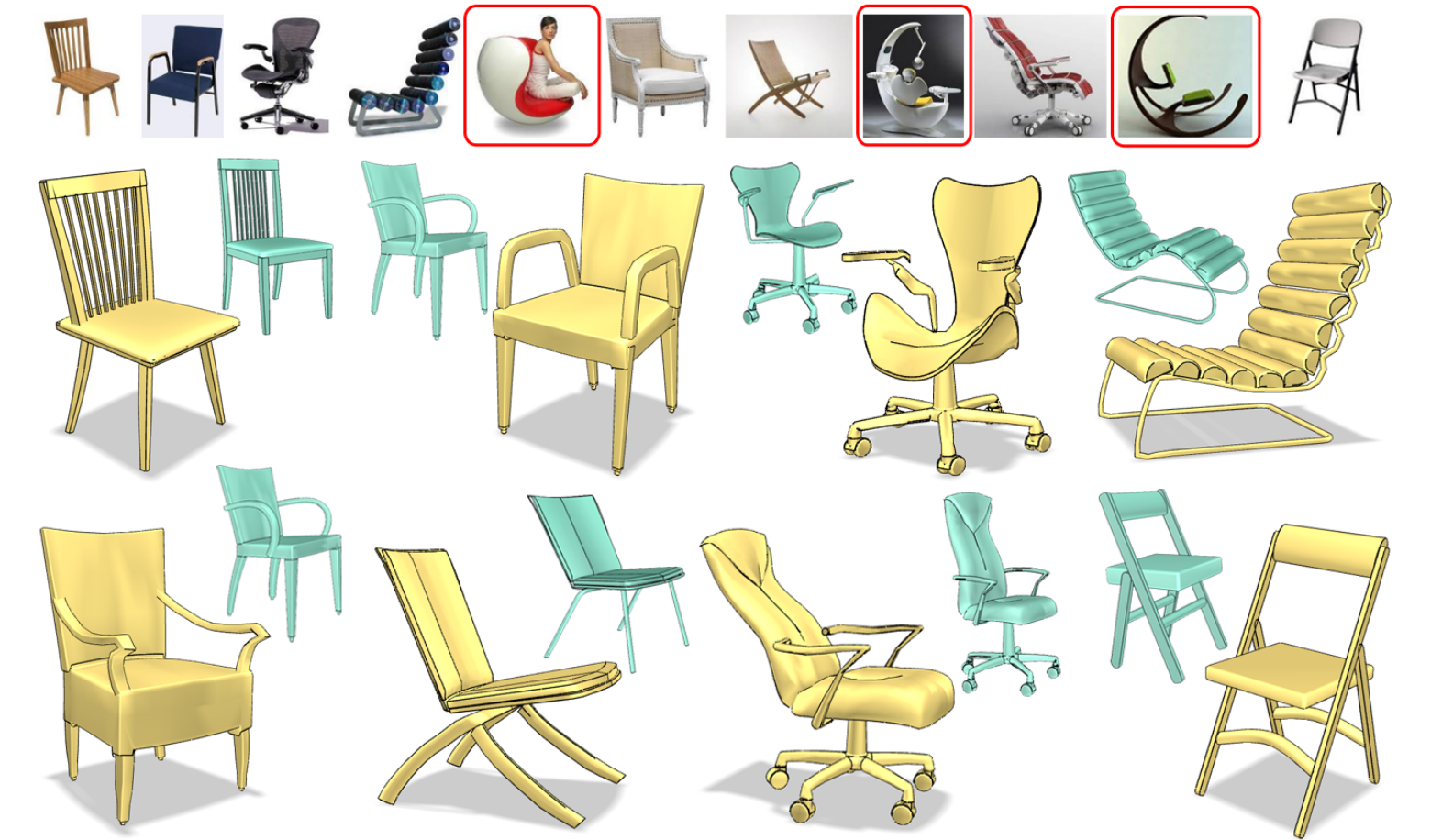,width=14cm}}
\caption{The chair images on the top are searched from Google image. The images in red boxes cannot be created by the system. The yellow images are 3D models created by the system and the green images are the representative candidate.}
\end{figure*}

\end{document}